\documentclass[10pt,twocolumn,letterpaper]{article}

\usepackage{cvpr}
\usepackage{times}
\usepackage{epsfig}
\usepackage{graphicx}
\usepackage{amsmath}
\usepackage{amssymb}
\usepackage{comment}
\usepackage{multirow}
\usepackage{booktabs}
\usepackage{amsmath,amssymb}
\usepackage{graphicx}
\usepackage{subfigure}
\usepackage{wrapfig}
\usepackage{color}

\usepackage[pagebackref=true,breaklinks=true,letterpaper=true,colorlinks,bookmarks=false]{hyperref}

\hypersetup{
    colorlinks=true,
    linkcolor=red,
    filecolor=magenta,      
    urlcolor=magenta,
}

\cvprfinalcopy 

\def\cvprPaperID{1418} 

\ifcvprfinal\fi
\begin{document}
	
	\title{Deep  Occlusion-Aware  Instance Segmentation with Overlapping BiLayers} 
	\author{
 Lei Ke$^1$\hspace{1.0cm}Yu-Wing Tai$^{2}$\hspace{1.0cm}Chi-Keung Tang$^1$
 \vspace{0.1cm}\\
 $^1$The Hong Kong University of Science and Technology\hspace{1.0cm}$^2$Kuaishou Technology
 \\
 \texttt{\footnotesize \{lkeab,cktang\}@cse.ust.hk, yuwing@gmail.com}}
	
	\maketitle
	
	\begin{abstract}	
		Segmenting highly-overlapping objects is challenging, because typically no distinction is made between real object contours and occlusion boundaries.
Unlike previous two-stage instance segmentation methods, we model image formation as composition of two overlapping layers, and propose~\textbf{B}ilayer~\textbf{C}onvolutional~\textbf{Net}work (\textbf{BCNet}), where the top GCN layer detects the occluding objects (occluder) and the bottom GCN layer infers partially occluded instance (occludee).
The explicit modeling of occlusion relationship with bilayer structure naturally decouples the boundaries of both the occluding and occluded instances, and considers the interaction between them during mask regression. 
We validate the efficacy of bilayer decoupling on both one-stage and two-stage object detectors with different backbones and network layer choices.
Despite its simplicity, extensive experiments on COCO and KINS show that our occlusion-aware BCNet achieves large and consistent performance gain especially for heavy occlusion cases.
Code is available at \url{https://github.com/lkeab/BCNet}.

	\end{abstract}
	
	\vspace{-2mm}
   \section{Introduction}
\footnotetext[1]{This research is supported in part by the Research Grant Council of the Hong Kong SAR under grant no. 16201420 and Kuaishou Technology.}

State-of-the-art approaches in instance segmentation often follow the Mask R-CNN~\cite{he2017mask} paradigm with the first stage detecting  bounding boxes, followed by  the second stage to segment instance masks. Mask R-CNN and its variants~\cite{liu2018path,cai2018cascade,chen2018masklab,huang2019mask,chen2019hybrid} have demonstrated notable performance, and most of the leading approaches in the COCO instance segmentation challenge~\cite{lin2014microsoft} have adopted this pipeline. However, we note that most incremental improvement comes from better backbone architecture designs, with little  attention paid in the instance mask regression after obtaining the ROI (Region-of-Interest) features from object detection. We observe that a lot of segmentation errors are caused by overlapping objects, especially for  object instances  belonging to the same class. 
This is because each instance mask is  individually regressed, and the regression process implicitly assumes the object in an ROI has almost complete contour, since most objects in the training data in COCO do not exhibit significant occlusions. 

\begin{figure}[!t]
	\centering
	\includegraphics[width=1.0\linewidth]{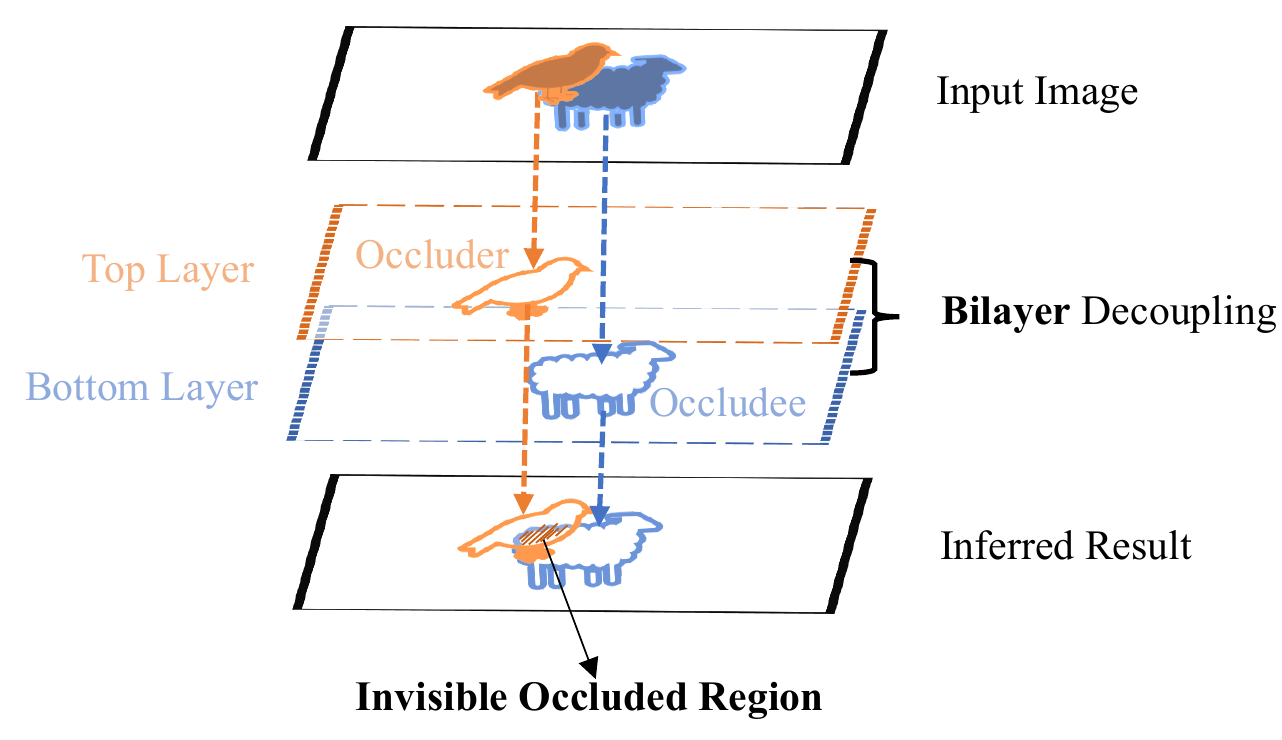}
	\caption{\textbf{Simplified illustration}. Unlike previous segmentation approaches operating on a single image layer (i.e., directly on the input image), we decouple overlapping objects into~\textit{two image layers}, where the top layer deals with the occluding objects (\textbf{occluder}) and the bottom layer for \textbf{occludee} (which is also referred to as target object in other methods as they do not explicitly consider the occluder). The overlapping parts of the two image layers indicate the invisible region of the occludee, which is explicitly modeled by our occlusion-aware BCNet framework.}
	\label{fig:teaser1}
	\vspace{-0.3cm}
\end{figure}

\begin{figure*}[!t]
	\centering
    \includegraphics[width=1.0\linewidth]{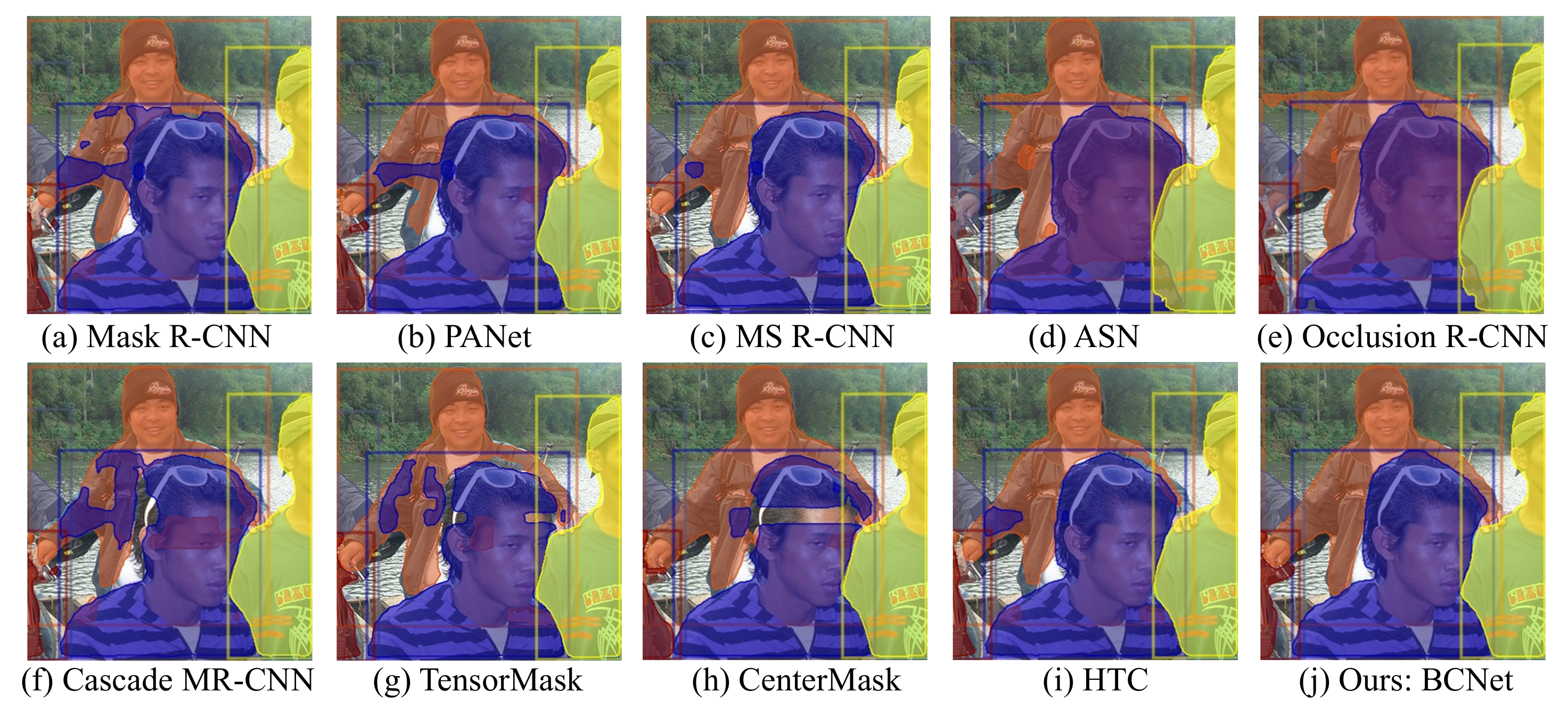}
	\vspace{-0.15in}
	\caption{Instance Segmentation on \textbf{COCO}~\cite{lin2014microsoft} validation set by a) Mask R-CNN~\cite{he2017mask}, b) PANet~\cite{liu2018path}, c) Mask Scoring R-CNN~\cite{huang2019mask}, d) ASN~\cite{qi2019amodal}, e) Occlusion R-CNN (ORCNN)~\cite{follmann2019learning}, f) Cascade Mask R-CNN~\cite{cai2018cascade}, g) TensorMask~\cite{chen2019tensormask}, h) CenterMask~\cite{lee2019centermask}, i) HTC~\cite{chen2019hybrid} and j) Our BCNet. Note that d) and e) are specially designed for amodal/occlusion mask prediction. In this example, the bounding box is given to compare the quality of different regressed instance masks.}
    \label{fig:fig1}
    \vspace{-0.1in}
\end{figure*}

We propose the Bilayer Convolutional Network (BCNet). As illustrated in Figure~\ref{fig:teaser1}, BCNet simultaneously regresses both occluding region (occluder) and partially occluded object (occludee) after ROI extraction, which  groups the pixels belonging to the occluding region and treat them equally as the pixels of the occluded object but in \textit{two separate image layers}, and thus naturally decouples the boundaries for both objects 
and considers the interaction between them during the mask regression stage.  

\begin{figure*}[!t]
	\centering
	\vspace{-0.2in}
	\includegraphics[width=1.0\linewidth]{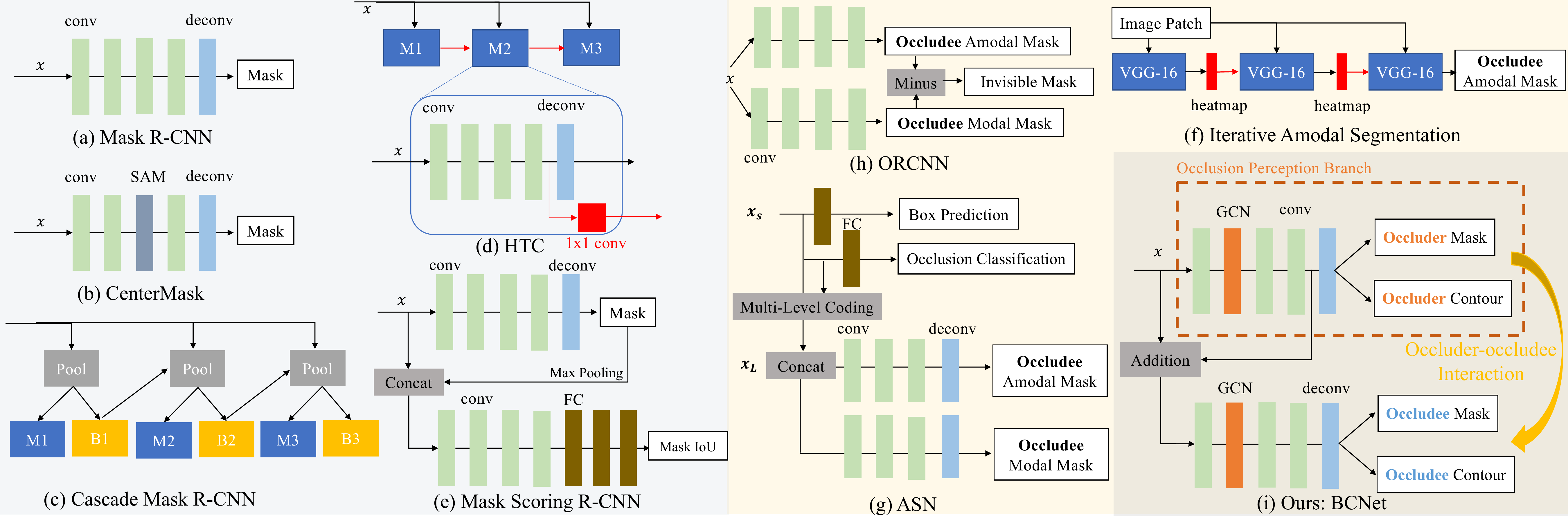}
	\vspace{-0.15in}
	\caption{A brief comparison of \textbf{mask head architectures}: a) Mask R-CNN~\cite{he2017mask}, b) CenterMask~\cite{lee2019centermask}, c) Cascade Mask R-CNN~\cite{cai2018cascade}, d) HTC~\cite{chen2019hybrid}, e) Mask Scoring R-CNN~\cite{huang2019mask}, f) Iterative Amodal Segmentation~\cite{li2016amodal}, g) ASN~\cite{qi2019amodal}, h) ORCNN~\cite{follmann2019learning}, where f), g) and h) are specially designed for amodal/occlusion mask prediction, i) Ours: BCNet. The input $\mathbf{x}$ denotes CNN feature after ROI extraction. {\tt Conv} is convolution layer with $3\times3$ kernel, {\tt FC} is the fully connected layer, {\tt SAM} is the spatial attention module. B$_t$ and M$_t$ respectively denote box and mask head at $t$-th stage. Unlike previous occlusion-aware mask heads, which only regress both modal and amodal masks from the occludee, our BCNet has a~\textit{bilayer GCN structure} and considers the \textbf{interactions between the top ``occluder" and bottom ``occludee"} in the same ROI. The \textbf{occlusion perception branch} explicitly models the occluding object by performing joint mask and contour predictions, and distills essential occlusion information for the second graph layer to segment target object (``occludee''). }
	\label{fig:motivation}
	\vspace{-0.1in}
\end{figure*}

Previous approaches resolve the mask conflict between neighboring objects through  non-maximum suppression or additional post-processing~\cite{liu2016multi,dai2016instance,li2016iterative,krahenbuhl2011efficient,hariharan2015hypercolumns}. Consequently, their results are over-smooth along boundaries or exhibit small gaps between neighboring objects. Furthermore, since the receptive field in the ROI observes multiple objects that belong to the same class, when the occluding regions were included as part of the occluded object, traditional mask head design falls short of resolving such conflict, leaving a large portion of error as shown in Figure~\ref{fig:fig1}.
We  compare BCNet with recent amodal segmentation methods~\cite{qi2019amodal,follmann2019learning}, which predict complete object masks, including the occluded region.
However, these amodal methods only regress single occluded target in the ROI, thus lacking  occluder-occludee interaction reasoning, making their specially designed decoupling structure suffer when handling mask conflict between highly-overlapping objects.
Correspondingly, 
Figure~\ref{fig:motivation} compares the architecture of our BCNet with previous mask head designs~\cite{he2017mask,liu2018path,huang2019mask,chen2019hybrid,lee2019centermask,cai2018cascade,qi2019amodal,follmann2019learning}. 

Our BCNet consists of two GCN layers with a cascaded structure, each respectively regresses the mask and boundaries of the occluding and partially occluded objects. We utilize GCN in our implementation because GCN can consider the non-local relationship between pixels, allowing for propagating information across pixels despite the presence of occluding regions. 
The explicit bilayer occluder-occludee relational modeling within the same ROI also makes our final segmentation results more explainable than previous methods. 
For object detector, we use the FCOS~\cite{tian2019fcos} owing to its efficient memory and running time, while noting that other state-of-the-art object detectors can also be used as demonstrated in our experiments.

Since our paper focuses on occlusion handling in instance segmentation, in addition to the original COCO evaluation, we extract a subset of COCO dataset containing both occluding objects and partially occluded objects to evaluate the robustness of our approach in comparison with other 
instance segmentation methods in occlusion handling. In this paper we also contribute the first large-scale occlusion aware instance segmentation datasets with ground-truth, complete object contours for {\em both} occluding and partially occluded objects. Extensive experiments show that our approach outperforms state-of-the-art methods in both the modal and amodal instance segmentation tasks.

   \section{Related Work}
   \paragraph{Instance Segmentation}
Two stage instance segmentation methods~\cite{li2017fully,he2017mask,liu2018path,chen2018masklab,cai2018cascade,chen2019hybrid,chen2019tensormask} achieve state-of-the-art performance by first detecting bounding boxes and then performing segmentation in each ROI region.
FCIS~\cite{li2017fully} introduces the position-sensitive score maps within instance proposals for mask segmentation. 
Mask R-CNN~\cite{he2017mask} extends Faster R-CNN~\cite{ren2015faster} with a FCN branch to segment objects in the detected box.
PANet~\cite{liu2018path} further integrates multi-level feature of FPN to enhance feature representation.
MS R-CNN~\cite{huang2019mask} mitigates the misalignment between mask quality and score.
CenterMask~\cite{lee2019centermask} is built upon the anchor free detector FCOS~\cite{tian2019fcos} with a SAG-Mask branch.
In contrast, our BCNet is a \textit{bilayer} mask prediction network for addressing the issues of heavy occlusion and overlapping objects in two-stage instance segmentation. Experiments validate that our approach leads to significant performance gain on {\em overall} instance segmentation performance not limited to heavily occluded cases. 
\vspace{-0.02in}

One-stage instance segmentation methods remove the bounding box detection and feature re-pooling steps.
AdaptIS~\cite{sofiiuk2019adaptis} produces masks for objects located on point proposals.
PolarMask~\cite{xie2019polarmask} models instance masks in  polar coordinates by instance center classification and dense distance regression.
YOLOACT~\cite{bolya2019yolact} introduces prototype masks with per-instance coefficients. 
SOLO~\cite{wang2019solo} applies the ``instance categories'' concept to directly output instance masks based on the location and size.
 Grouping-based approaches~\cite{kirillov2017instancecut,arnab2017pixelwise,liu2017sgn,liu2018affinity,bai2017deep,kong2018recurrent}  regard  segmentation as a bottom-up grouping task by first producing pixel-wise predictions followed by grouping object instances in the post-processing stage.
These one-stage methods, with simpler procedures than their two-stage counterparts, are more efficient but tend to be less accurate.
\vspace{-0.2in}
\paragraph{Occlusion Handling} 
Methods for occlusion handling have been proposed~\cite{sun2005symmetric,winn2006layout,gao2011segmentation,chen2015parsing,yang2011layered,hsiao2014occlusion,gao2011segmentation,zhu2017semantic,yan2019visualizing}.
A layout consistent random field is used in~\cite{winn2006layout} to segment images of cars and faces by imposing asymmetric local spatial constraints.
Ghiasi \textit{et al}.~\cite{ghiasi2014parsing} model occlusion by learning deformable models with local templates for human pose estimation while~\cite{Ke_2020_ECCV} reconstructs dense 3D shape for vehicle pose.
Tighe \textit{et al}.~\cite{tighe2014scene} build a histogram to predict occlusion overlap scores between two classes for inferring occlusion order in the scene parsing task.
Chen \textit{et al}.~\cite{chen2015multi} handle occlusion by incorporating category specific reasoning and exemplar-based shape prediction for instance segmentation.
For pedestrian detection with occlusion, bi-box regression is proposed in~\cite{zhou2018bi} for both full body and visible part estimation while repulsion loss~\cite{wang2018repulsion} and aggregation loss~\cite{zhang2018occlusion} are designed to improve the detection accuracy.
SeGAN~\cite{ehsani2018segan} learns occlusion patterns by segmenting and generating the invisible part of an object.
Recently, OCFusion~\cite{lazarow2019learning} uses an additional branch to model instances fusion process for replacing detection confidence in panoptic segmentation.
A self-supervised scene de-occlusion method is proposed in~\cite{zhan2020self} by recovering the occlusion ordering and completing the mask and content for the invisible object parts.

Compared to these methods, our BCNet  tackles occlusion by explicitly modeling occlusion patterns in shape and appearance. This equips the segmentation model  with strong occlusion perception and reasoning capability. Our bi-layer approach can be smoothly integrated into state-of-the-art segmentation framework for end-to-end training.
\vspace{-0.2in}

\paragraph{Amodal Instance Segmentation} 
Different from traditional segmentation which only focuses on visible regions, amodal instance segmentation can predict the occluded parts of object instances. Li and Malik~\cite{li2016amodal} first propose a method by extending~\cite{li2016iterative}, which iteratively enlarges the modal bounding box following the direction of high heatmap values and synthetically adds occlusion. Zhu~\textit{et al}.~\cite{zhu2017semantic} propose a COCO amodal dataset with 5000 images from the original COCO and use AmodalMask as a baseline, which is SharpMask~\cite{pinheiro2016learning} trained on amodal ground truth. COCOA~\textit{cls}~\cite{follmann2019learning} augments this dataset by assigning class-labels to the objects while SAIL-VOS dataset  in~\cite{hu2019sail} is targeted for video object segmentation. In autonomous driving, Qi~\textit{et al.}~\cite{qi2019amodal} establish the large-scale KITTI~\cite{geiger2012we} InStance segmentation dataset (KINS) and present ASN to improve  amodal segmentation performance.

Comparing to  most of the amodal and occlusion reasoning methods which regress single occluded object boundary directly on the input (single-layered) image, our BCNet decouples overlapping objects in the same ROI into two disjoint graph layers by predicting the complete object segments (Figure~\ref{fig:teaser1}), 
where the occludee is segmented under the guidance from the shape and location of the occluder.
   
   \section{Occlusion-Aware Instance Segmentation}

\begin{figure*}[!t]
	\centering
	\vspace{-0.2in}
	\includegraphics[width=0.9\linewidth]{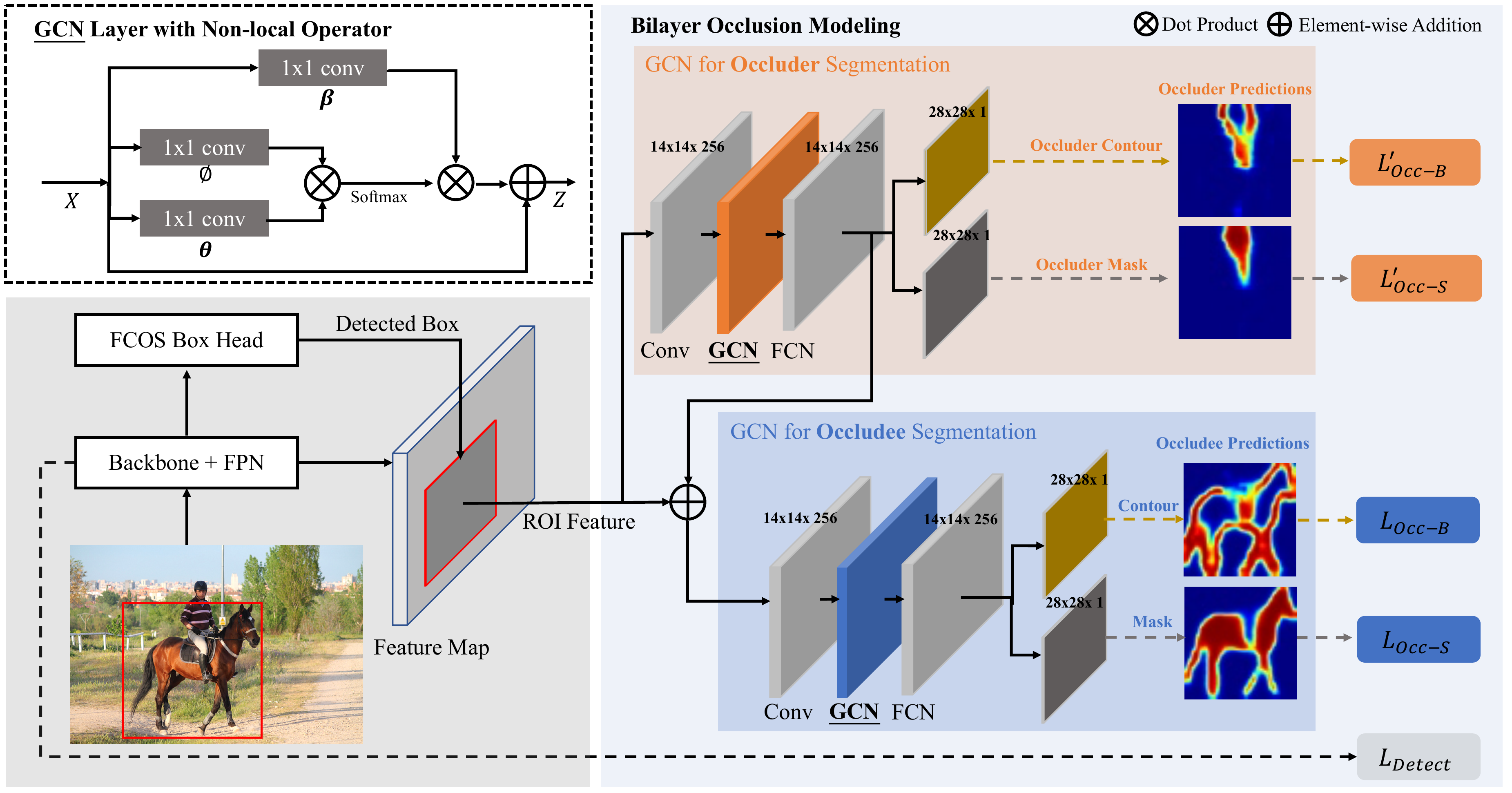}
	\vspace{-0.05in}
	\caption{Architecture of our BCNet with bilayer occluder-occludee relational modeling, which consists of three modules; (1)~Backbone~\cite{he2016deep} with FPN for feature extraction from input image; (2)~Detection branch~\cite{tian2019fcos} for predicting instance proposals; (3)~BCNet with bilayer GCN structure for mask prediction. For cropped ROI feature, the first GCN explicitly models occluding regions (occluder) by simultaneously detecting occlusion contours and masks, which distills essential shape and position information to guide the second GCN in mask prediction for the occludee. We utilize the non-local operator~\cite{wang2018non, wang2018videos} detailed in section~\ref{BCNet} to implement the GCN layer. Visualization results are resized to square size.}
	\label{fig:framework}
	\vspace{-0.2in}
\end{figure*}

We first give an overview to the overall instance segmentation framework, and then describe the proposed Bilayer Graph Convolutional Network (BCNet) with explicit occluder-occludee modeling. Finally, we specify the objective functions for the whole network optimization, and provide details of training and inference process.

\subsection{Overview}

\paragraph{Motivation}
For images with heavy occlusion, multiple overlapping objects in the same bounding box may result in confusing instance contours from both real objects and occlusion boundaries. 
The mask head design of Mask R-CNN and its variants~\cite{huang2019mask,chen2019hybrid,cai2018cascade,qi2019amodal,follmann2019learning} in Figure~\ref{fig:motivation} directly regress the occludee with a fully convolutional network, which neglects both the occluding instances and the overlapping relations between objects. 
To mitigate this limitation, BCNet extends existing two stage instance segmentation methods, by adding an occlusion perception branch parallel to the traditional target prediction pipeline. Thus, the interactions between objects within the ROI region can be well considered during the mask regression stage.

Figure~\ref{fig:framework} gives the overall~\textbf{architecture} of BCNet for addressing occlusion in instance segmentation.
Following typical models~\cite{he2017mask,lee2019centermask} for instance segmentation, our model has  three parts:
(1) Backbone~\cite{he2016deep} with FPN~\cite{lin2017feature} for ROI feature extraction;
(2) Object detection head in charge of predicting bounding boxes
as instance proposals. We employ FCOS~\cite{tian2019fcos} as the object detector owing to its anchor-free efficiency though our method is flexible and can deploy any existing fully supervised object detectors~\cite{ren2015faster,redmon2016you,lin2017focal};
(3) The \textit{occlusion-aware mask head}, BCNet, uses bilayer GCN structure for decoupling overlapping relations and segments the instance proposals obtained from the object detection branch.
BCNet reformulates the traditional class-agnostic segmentation as two complementary tasks: occluder modeling using the first GCN and occludee prediction with the second GCN, where the auxiliary predictions from the first GCN provide rich occlusion cues, such as shape and positions of occluding regions, to guide target (occludee) object segmentation.

\paragraph{Work Flow} Given an input image, the backbone network equipped with FPN first extracts intermediate convolutional features for downstream processing. 
Then, the object detection head predicts bounding boxes with positions as well as categories for potential instances, and prepares the cropped ROI feature for BCNet to produce segmentation masks.
The occlusion perception branch consists of the first GCN layer followed by FCN (two convolution layers), which is targeted for modeling occluding regions by jointly detecting contours and masks. 
Forming a residual connection, the distilled occlusion feature is element-wise added to the original input ROI feature and passed to second GCN.
Finally, the second GCN, which has a similar structure to the first GCN, segments the occludee guided by this occlusion-aware feature and outputs contours and masks for the partially occluded instance.

\subsection{Bilayer Occluder-Occludee  Modeling}
\label{BCNet}
\paragraph{Bilayer GCN Structure for Instance Segmentation} 
Recently, Graph Convolutional Network (GCN)~\cite{kipf2017semi} has been  adopted to model long-range relationships in images~\cite{chen2019graph,zhang2019dual,li2018beyond} and videos~\cite{wang2018videos}. 
Given highly-overlapping objects, pixels belonging to the same partially occluded object may be separated into disjoint subregions by the occluder.
Thus, we adopt GCN as our basic block due to its non-local property~\cite{wang2018non}, where each graph node represents a single pixel on the feature map.
To explicitly model the occluding region, we further extend the single GCN block to the bilayer GCN structure as shown in Figure~\ref{fig:framework}, which constructs two orthogonal graphs in a single general framework. 

Following~\cite{wang2018videos}, given an adjacency graph~$\mathcal{G=\langle\mathcal{V}, \mathcal{E}}\rangle$ with edges~$\mathcal{E}$ among nodes~$\mathcal{V}$, we represent the graph convolution operation as,
\begin{equation}
\mathbf{Z} = \mathbf{\sigma }(\mathbf{A}\mathbf{X}\mathbf{W}_g) + \mathbf{X},
\end{equation} 
where~$\mathbf{X} \in \mathbb{R}^{N\times K}$ is the input feature, $N=H\times W$ is the number of pixel grids within the ROI region and $K$ is the feature dimension for each node, $\mathbf{A} \in \mathbb{R}^{N\times N}$ is the adjacency matrix for defining neighboring relations of graph nodes by feature similarities, and $\mathbf{W}_g \in \mathbb{R}^{K\times K'}$ is the learnable weight matrix for the output transform, where~$K'=K$ in our case. The output feature $\mathbf{Z} \in \mathbb{R}^{N\times K'}$ consists of the updated node feature by global information propagation within the whole graph layer, which is obtained after non-linear functions~$\mathbf{\sigma }(\cdot)$ including layer normalization~\cite{ba2016layer} and ReLU functions. We add a residual connection after the GCN layer.

To construct the adjacency matrix~$\mathbf{A}$, we define the pairwise similarity between every two graph nodes $\mathbf{x}_i, \mathbf{x}_j$ by dot product similarity as,
\begin{equation}
\mathbf{A}_{ij} = \mathit{softmax}(F(\mathbf{x}_i, \mathbf{x}_j)), 
\end{equation}
\begin{equation}
F(\mathbf{x}_i,\mathbf{x}_j) = \theta(\mathbf{x}_i)^{T}\phi(\mathbf{x}_j),
\end{equation}
where $\theta$ and $\phi$ are two trainable transformation function implemented by $1\times1$ convolution as shown in the non-local operator part of Figure~\ref{fig:framework}, so that high confidence edge between two nodes corresponds to larger feature similarity.

In our bilayer GCN structure, we further define $\mathcal{G}^i$ to indicate the $i$th graph, $X_{roi}$ for the input ROI feature and~$\mathbf{W}_f$ for weights in FCN layers, then the complete formulae are:
\begin{equation}
\mathbf{Z}^1 = \mathbf{\sigma }(\mathbf{A}^1\mathbf{X}_f\mathbf{W}_g^{1}) + \mathbf{X}_{f}, 
\end{equation} 
\begin{equation}
\mathbf{X}_f = \mathbf{Z}^0\mathbf{W}_{f}^0 + \mathbf{X}_{roi}, 
\end{equation}
\begin{equation}
\mathbf{Z}^0 = \mathbf{\sigma}(\mathbf{A}^0\mathbf{X}_{roi}\mathbf{W}_{g}^0)+\mathbf{X}_{roi}.
\end{equation} 
For connecting the two GCN blocks, the output feature $\mathbf{Z}^0$ of the occluder from the first GCN is directly added to~$\mathbf{X}_{roi}$ to obtain the fused~\textit{occlusion-aware} feature~$\mathbf{X}_f$, which is the input for the second GCN layer to output~$\mathbf{Z}^1$ for occludee mask prediction.

Compared to previous class-agnostic mask head with single layer structure, where there is only binary label (foreground/background) per pixel,  the bilayer GCN additionally constructs a new semantic graph space for~\textit{occluding region}. Thus a pixel node in overlapping areas in ROI can concurrently correspond to two different states in bilayer graph. 
While other choices may exist, we believe modeling GCN as a dual-layered structure as shown in Figure~\ref{fig:framework} is a natural choice for handling occlusion.

\paragraph{Occluder-occludee Modeling}
We explicitly model occlusion patterns by detecting both contours and masks for the occluders using the first GCN layer.
Since the second GCN layer jointly predicts contours for the occludee, the overlap between the two layers can be directly identified as occlusion boundary which can thus be distinguished from  real object contour (e.g., the occluder and occludee prediction on the rightmost of Figure~\ref{fig:framework}).
The rationale behind this design is that such irregular occlusion boundary unrelated to the occludee is confusing, which in turn provides essential cues for decoupling occlusion relations. Besides, accurate boundary localization  explicitly contributes to segmentation mask prediction.

The module for occluder modeling is designed in a simple yet effective way: one 3$\times$3 convolutional layer followed by one GCN layer and one FCN layer. Then we feed the output to the up-sampling layer and one 1$\times$1 convolutional layer to obtain one channel feature map for joint boundary and mask predictions.
The boundary detection for occluder is trained with loss~$\mathcal{L^\prime}_{\text{Occ-B}}$:
\begin{equation}
\mathcal{L^\prime}_\text{Occ-B} = \mathcal{L}_{\text{BCE}}(W_B\mathcal{F}_{occ}(\mathbf{X}_{roi}), \mathcal{GT}_{B}),
\label{eq:eq4}
\end{equation}
where $\mathcal{L}_{\text{BCE}}$ denotes the binary cross-entropy loss, $\mathcal{F}_{occ}$ denotes the nonlinear transformation function of the occlusion modeling module, $W_{B}$ is the boundary predictor weight, $\mathbf{X}_{roi}$ is the cropped FPN feature map given by RoIAlign operation for the target region, and $\mathcal{GT}_{B}$ is the off-the-shelf occluder boundary that can be readily computed from mask annotations.

For occluder mask prediction, it utilizes the shared feature $\mathcal{F}_{occ}(\mathbf{X}_{roi})$, which is jointly optimized by boundary prediction. The segmentation loss $\mathcal{L^\prime}_{\text{Occ-S}}$ for occluder modeling is designed as
\begin{equation}
\mathcal{L^\prime}_{\text{Occ-S}} = \mathcal{L}_{\text{BCE}}(W_S\mathcal{F}_{occ}(\mathbf{X}_{roi}), \mathcal{GT}_{S}),
\label{eq:eq5}
\end{equation}
where $W_S$ denotes the trainable weight of segmentation mask predictor by $1\times1$ convolutional layer, and $\mathcal{GT}_{S}$ is the mask annotations for the occluder.

\subsection{End-to-end Parameter Learning}
The whole instance segmentation framework can be trained in an end-to-end manner defined by a multi-task loss function $\mathcal{L}$ as,
\begin{equation}
\mathcal{L} = \lambda_1\mathcal{L}_{\text{Detect}} + \mathcal{L}_{\text{Occluder}} + \mathcal{L}_{\text{Occludee}},
\end{equation}
\vspace{-0.2in}
\begin{equation}
\mathcal{L}_{\text{Occluder}} = \lambda_2\mathcal{L^\prime}_{\text{Occ-B}} + \lambda_3\mathcal{L^\prime}_{\text{Occ-S}}
\end{equation}
\begin{equation}
 \mathcal{L}_{\text{Occludee}} = \lambda_4\mathcal{L}_{\text{Occ-B}} + \lambda_5\mathcal{L}_{\text{Occ-S}},
\end{equation}
where~$\mathcal{L}_{\text{Occ-B}}$ and $\mathcal{L}_{\text{Occ-S}}$ denote respectively the boundary detection and mask segmentation losses in the second GCN layer for the occludee, which are similar to Eq.~\ref{eq:eq4} and Eq.~\ref{eq:eq5}.~$\mathcal{L}_{\text{Detect}}$ supervises  both the position prediction and the category classification borrowed from the FCOS~\cite{tian2019fcos} detector, 
\begin{equation}
\mathcal{L}_{\text{Detect}} = \mathcal{L}_{\text{Regression}} + \mathcal{L}_{\text{Centerness}} + \mathcal{L}_{\text{Class}},
\end{equation}
and $\lambda_1$, $\lambda_2$, $\lambda_3$, $\lambda_4$ and $\lambda_5$ are hyper-parameter weights to balance the loss functions, which are tuned to be $\{1, 0.5, 0.25, 0.5, 1.0\}$ respectively on the validation set. 

\paragraph{Training:}
For training the first GCN layer of BCNet, since partial occlusion cases only occupy a small fraction compared to the complete objects in COCO, we filter out part of the non-occluded ROI proposals to keep occlusion cases taking up 50\% for balance sampling. SGD with momentum is employed for training 90K iterations which starts with 1K constant warm-up iterations. The batch size is set to 16 and initial learning rate is 0.01.  In ablation study, ResNet-50-FPN~\cite{he2016deep} is used as backbone and the input images are resized without changing the aspect ratio by keeping the shorter side and longer side of no more than 600 and 900 pixels respectively. For leaderboard comparison, we adopt the scale-jitter where the shorter image side is randomly sampled from [640, 800] following~\cite{lee2019centermask,chen2019tensormask,bolya2019yolact}. 

\paragraph{Inference:} 
During inference, the mask head predicts masks for the occluded target object in the high-score box proposals (no more than 50) generated by the FCOS detector, where the first GCN layer only produces occlusion-aware feature as input for the second GCN. 
   
	
	\section{Experiments}
	\subsection{Experimental Setup}

\paragraph{COCO and COCO-OCC}
We conduct experiments on COCO dataset~\cite{lin2014microsoft}, where we train on 2017{\it train} (115k images) and evaluate results on both 2017{\it val} and 2017{\it test-dev} using the standard metrics. For further investigating segmentation performance with occlusion handling, we propose a subset split, called COCO-OCC, which contains 1,005 images extracted from the validation set (5k images) where the overlapping ratio between the bounding boxes of objects is at least 0.2. Segmenting COCO-OCC with highly overlapping objects is much more difficult than 2017{\it val}, where we observe a performance gap around 3.0$AP$ for the same model in the experiment section.

\paragraph{KINS and COCOA}
We also evaluate BCNet on two amodal instance segmentation benchmarks: 
(1) \textbf{KINS}~\cite{qi2019amodal}, built on the original KITTI~\cite{geiger2012we}, is the largest amodal segmentation benchmark for traffic scenes with both annotated amodal and modal masks for instances. BCNet is trained on the training split (7,474 images and 95,311 instances) and tested on the testing split (7,517 images and 92,492 instances) following the setting in~\cite{qi2019amodal}.
(2) \textbf{COCOA}~\cite{zhu2017semantic} is a subpart of COCO~\cite{lin2014microsoft}, where we train BCNet on the official training split (2,500 images) and test on the validation split (1,323 images). Note that each instance has no class label and we only use the modal and amodal mask labels for the COCOA dataset.

\paragraph{Synthetic Occlusion Dataset}
Since most objects in COCO do not exhibit significant occlusions, we synthesize a large-scale instance segmentation dataset which contains 100k  images following uniform class distribution for instances among the 80 categories in COCO. Each synthetic image has \textit{true and complete} object contours for {\em both} occluding and partially occluded objects, thus allowing the explicit modeling of occlusion relationship between the occlusion regions and occluded objects. On the other hand, COCOA~\cite{zhu2017semantic}, which has only 5,000 images, relies on user annotation on a given training image for ``guessing" occluded object boundaries. More details on our occlusion dataset synthesis process are provided in the supplementary file.  

\subsection{Ablation Study}

\paragraph{Effect of Explicit Occlusion Modeling} We validate the efficacy of different components proposed for explicit occlusion modeling on the first GCN layer. Table~\ref{tab:fist_GCN} tabulates the  quantitative comparison: 1) Baseline: BCNet with no explicit occlusion modeling targets; 2) modeling segmentation masks for occluding regions (\textbf{occluder}); 3)  modeling contours of the occluding regions; 4) \textbf{joint} occlusion modeling on both masks and contours.   Compared to the baseline, joint occlusion modeling produces the most obvious improvement especially for the heavy occlusion cases, which promotes mask $AP$ on the standard validation set from 32.65 to 33.43, and the $AP$ on the proposed COCO-OCC split is increased from 29.04 to 30.37.

\begin{table}[!h]
	\caption{Effect of the first GCN for  occlusion modeling by predicting contours and masks on COCO with ResNet-50-FPN model.}
	\centering
	\resizebox{0.85\linewidth}{!}{
		\begin{tabular}{c | c | c | c | c | c }
			\toprule
			\multicolumn{2}{c|}{Occlusion (Occluder) Modeling} & \multicolumn{2}{c|}{COCO-OCC} & \multicolumn{2}{c}{COCO} \\
			\cline{1-6} 
			Contour & Mask & $AP$ & $AP_{50}$ & $AP$ & $AP_{50}$\\
			\midrule
			&  & 29.04 & 49.22 & 32.65 & 52.39 \\
			& \checkmark & 29.65 & 49.42 & 33.25 & 52.82 \\
			\checkmark  & & 30.18 & 49.94 & 33.41 & 53.02 \\
			\checkmark  & \checkmark  & \textbf{30.37} & \textbf{50.40} & \textbf{33.43} &  \textbf{53.12} \\
			\bottomrule
		\end{tabular}
	}
	\vspace{-0.2in}
	\label{tab:fist_GCN}
\end{table}

\paragraph{Effect of Bilayer Occluder-occludee Modeling} Built on the first GCN layer with explicit occlusion modeling, we further validate the second GCN layer in Table~\ref{tab:second_GCN}, which demonstrates the importance of~\textit{occlusion-aware} feature \textit{guidance} for the second GCN layer to segment target object (\textbf{occludee}) by boosting 1.23 $AP$ on COCO-OCC, and 1.06 $AP$ on COCO respectively. Table~\ref{tab:bilayer} shows the results comparison on adopting the proposed~\textit{bilayer structure} and existing direct regression model with single layer. On the COCO-OCC split, bilayer GCN improves $AP$ from 29.63 to 30.68 compared to single GCN, and bilayer FCN boosts the performance of single FCN from 28.43 to 30.12.

\begin{table}[!h]
	\caption{Effect of the second GCN for detecting occludee contours for final mask prediction \textbf{\textit{guided}} by the output of first GCN.}
	\centering
	\resizebox{0.95\linewidth}{!}{
		\begin{tabular}{c | c | c | c | c | c | c }
			\toprule
			\multicolumn{3}{c|}{Target (Occludee) Modeling} & \multicolumn{2}{c|}{COCO-OCC} & \multicolumn{2}{c}{COCO} \\
			\cline{1-7} 
			Guidance & Contour & Mask & $AP$ & $AP_{50}$ & $AP$ & $AP_{50}$\\
			\midrule
			& & \checkmark & 29.45 & 49.73 & 32.56 & 52.21 \\
			\checkmark & & \checkmark & 30.37 & 50.40 & 33.43 & 53.12 \\
			\checkmark & \checkmark  & \checkmark  & \textbf{30.68} & \textbf{50.62} & \textbf{33.62} & \textbf{53.26} \\
			\bottomrule
		\end{tabular}
	}
	\vspace{-0.2in}
	\label{tab:second_GCN}
\end{table}

\vspace{-0.1in}
\paragraph{Using FCN or GCN?} Table~\ref{tab:bilayer} also reveals the advantage of GCN over FCN, where GCN achieves consistent superior performance both in the singe layer and bilayer structure.
We also compute the number of parameters of each model and find that although GCN has more trainable parameters, the increased model size is acceptable compared to performance gain, because the feature size of input ROI has been down-sampled to only 14$\times$14 (spatial size) with 256 channels.

\begin{table}[!h]
	\caption{Effect of~\textbf{bilayer structure} using \textbf{GCN vs. FCN} implementation.}
	\centering
	\resizebox{0.95\linewidth}{!}{
		\begin{tabular}{c | c | c | c | c | c | c | c}
			\toprule
			\multirow{2}{*}{Structure} & \multirow{2}{*}{FCN} & \multirow{2}{*}{GCN} & \multicolumn{2}{c|}{COCO-OCC} & \multicolumn{2}{c|}{COCO} & \multirow{2}{*}{Params}\\
			\cline{4-7} 
			& & & $AP$ & $AP_{50}$ & $AP$ & $AP_{50}$ &\\
			\midrule
			\multirow{2}{*}{Single Layer} & \checkmark  & & 28.43 & 48.24 & 33.01 & 52.62 & 51.0M\\
			& & \checkmark & 29.63 & 49.59 & 33.14 & 52.81 & 51.4M\\
			\midrule
			\multirow{2}{*}{Bilayer} & \checkmark  & & 30.12 & 49.04 & 33.16 & 52.80 & 53.4M\\
			& & \checkmark & \textbf{30.68} & \textbf{50.62} & \textbf{33.62} & \textbf{53.26} & 54.0M\\
			\bottomrule
		\end{tabular}
	}
	\vspace{-0.3in}
	\label{tab:bilayer}
\end{table}

\paragraph{Influence of Object Detector} 
To investigate the influence of object detectors to BCNet, besides using one-stage detector FCOS~\cite{tian2019fcos}, we also use representative two-stage detector Faster R-CNN~\cite{ren2015faster} to perform experiments.  As shown in Table~\ref{tab:detector}, the performance gain brought by BCNet is consistent, with an improvement of 2.23 (for FCOS) and 2.04 (for Faster R-CNN) mask $AP$ on COCO-OCC respectively. Here, baseline denotes mask head design in Mask R-CNN.

\begin{table}[!h]
	\caption{Influence of the object detector (FCOS vs. Faster R-CNN) on BCNet.}
	\centering
	\resizebox{0.95\linewidth}{!}{
		\begin{tabular}{ l | c | c | c | c | c}
			\toprule
			\multirow{2}{*}{Model} & \multicolumn{2}{c|}{COCO-OCC} & \multicolumn{2}{c|}{COCO} & \multirow{2}{*}{Params} \\
			\cline{2-5} 
			& $AP$ & $AP_{50}$ & $AP$ & $AP_{50}$ \\
			\midrule
			FCOS~\cite{lee2019centermask} + Baseline  & 28.43 & 48.24 & 33.01 & 52.62 & 51.0M\\
			FCOS~\cite{tian2019fcos} + Ours & \textbf{30.68} & \textbf{50.62} & \textbf{33.62} & \textbf{53.26} & 54.0M\\
			\midrule
			Faster R-CNN~\cite{he2017mask} + Baseline & 29.67 & 49.95 & 33.45 & 53.70 & 60.0M\\
			Faster R-CNN~\cite{ren2015faster} + Ours & \textbf{31.71} & \textbf{51.15} & \textbf{34.61} & \textbf{54.41} & 63.2M \\
			\bottomrule
		\end{tabular}
	}
	\vspace{-0.15in}
	\label{tab:detector}
\end{table}

\subsection{Performance Comparison and Analysis}

\paragraph{Comparison with SOTA Methods}
Table~\ref{table:fully} compares BCNet with  state-of-the-art instance segmentation methods on COCO dataset.
BCGN achieves consistent improvement on different backbones and object detectors, demonstrating its effectiveness by outperforming both PANet~\cite{liu2018path} and Mask Scoring R-CNN~\cite{huang2019mask} by 1.5  \textit{AP} using Faster R-CNN, and exceeding CenterMask~\cite{lee2019centermask} by 1.3 \textit{AP} using FCOS. Our single model achieves comparable result with HTC~\cite{chen2019hybrid}, which uses a 3-stage cascade refinement with multiple object detectors and mask heads, and far more parameters.

\vspace{-0.2in}
\begin{table*}[h]
	\begin{minipage}[t]{0.33\linewidth}
		\caption{Results on the COCOA dataset.}
		\centering
		\resizebox{1.0\linewidth}{!}{
			\begin{tabular}{l | c | c | c}
				\toprule
				Model & $AP_{all}$ & $AP_{t}$ & $AP_{s}$\\
				\midrule
				AmodalMask~\cite{zhu2017semantic} & 5.7 & 5.9 & 0.8 \\
				AmodalMRCNN~\cite{follmann2019learning} & 21.51 & 21.09 & 9.0 \\
				ORCNN~\cite{follmann2019learning} & 20.32 & 20.63 & 7.8 \\
				\midrule
				\textbf{BCNet} & \textbf{23.09} & \textbf{22.72} & \textbf{9.53} \\
				\bottomrule
			\end{tabular}
		}
		\label{tab:cocoa}
	\end{minipage}
	\hfill
	\begin{minipage}[t]{0.33\linewidth}
		\caption{Results on the KINS dataset.}
		\centering
		\resizebox{0.85\linewidth}{!}{
			\begin{tabular}{l | c | c }
				\toprule
				Model & $AP_{Det}$ & $AP_{Seg}$ \\
				\midrule
				Mask R-CNN~\cite{follmann2019learning} & 26.97 & 24.93 \\
				Mask R-CNN + ASN~\cite{qi2019amodal} & 27.86 & 25.62 \\
				PANet~\cite{liu2018path} & 27.39 & 25.99 \\
				PANet + ASN~\cite{qi2019amodal} & 28.41 & 26.81 \\
				\midrule
				\textbf{BCNet} & \textbf{28.87} & \textbf{27.30} \\
				\bottomrule
			\end{tabular}
		}
		\label{tab:kins}
	\end{minipage}
	\hfill
	\begin{minipage}[t]{0.33\linewidth}
		\caption{Results on COCO-OCC split.}
		\centering
		\resizebox{0.65\linewidth}{!}{
			\begin{tabular}{l | c | c }
				\toprule
				Model & $AP$ & $AP_{50}$ \\
				\midrule
				Mask R-CNN~\cite{he2016deep} & 29.67 & 49.95 \\
				CenterMask~\cite{lee2019centermask} & 29.05 & 49.07 \\
				MS R-CNN~\cite{huang2019mask} & 30.32 & 50.01 \\
				\midrule
				\textbf{Ours} & 31.71 & 51.15\\
				\textbf{Ours + Synthetic} & \textbf{32.89} & \textbf{53.25}\\
				\bottomrule
			\end{tabular}
		}
		\label{tab:occ_split}
	\end{minipage}
	\vspace{-0.1in}
\end{table*}

\begin{table*}[!t]
	\caption{Comparison with SOTA methods on COCO {\it test-dev} set. The mask AP is reported and all entries are single-model results. Note that HTC~\cite{chen2019hybrid} adopts 3-stage cascade refinement with multiple object detectors and mask heads. All methods are trained on COCO \emph{train2017}.} 
	\vspace{-0.05in}
	\begin{center}{\small
			\resizebox{0.8\linewidth}{!}{
				\begin{tabular}{cc|c|ccc|ccc}
					\hline
					& Method & Backbone & $AP$ & $AP_{50}$ & $AP_{75}$ & $AP_{S}$ & $AP_{M}$ & $AP_{L}$ \\ 
					\hline
					\multirow{4}{*}{} & Mask R-CNN~\cite{he2017mask} & ResNet-50 & 35.6 & 57.6 & 38.1 & 18.7 & 38.3 & 46.6 \\ 
					& PANet~\cite{liu2018path} & ResNet-50 & 36.6 & 58.0 & 39.3 & 16.3 & 38.1 & \textbf{52.4} \\
					& \textbf{BCNet + Faster R-CNN~\cite{ren2015faster}} & ResNet-50 & \textbf{38.4} & \textbf{59.6} & \textbf{41.5} & \textbf{21.9} & \textbf{40.9} & 49.3 \\
					\hline
					& Mask R-CNN~\cite{he2017mask} & ResNet-101 & 37.0 & 59.2 & 39.5 & 17.1 & 39.3 & 52.9 \\ 
					& MaskLab~\cite{chen2018masklab} & ResNet-101 & 37.3 & 59.8 & 39.6 & 19.1 & 40.5 & 50.6 \\
					& Mask Scoring R-CNN~\cite{huang2019mask} & ResNet-101 & 38.3 & 58.8 & 41.5 & 17.8 & 40.4 & 54.4 \\
					& BMask R-CNN~\cite{ChengWHL20} & ResNet-101 & 37.7 & 59.3 & 40.6 & 16.8 & 39.9 & \textbf{54.6} \\
					& HTC~\cite{chen2019hybrid} & ResNet-101 & 39.7 & \textbf{61.8} & 43.1 & 21.0 & 42.2 & 53.5 \\
					& \textbf{BCNet + Faster R-CNN~\cite{ren2015faster}}      & ResNet-101 & {\bf 39.8}    & 61.5 & {\bf 43.1} & {\bf 22.7} & {\bf 42.4} & 51.1  \\
					\hline
					\hline
					\multirow{5}{*}{}& YOLACT~\cite{bolya2019yolact}       & ResNet-101 & 31.2 & 50.6 & 32.8 & 12.1 & 33.3 & 47.1 \\
					& TensorMask~\cite{chen2019tensormask}   & ResNet-101 & 37.1 & 59.3 & 39.4 & 17.4 & 39.1 & 51.6 \\
					& ShapeMask~\cite{kuo2019shapemask}      & ResNet-101 & 37.4 & 58.1 & 40.0 & 16.1 & 40.1 & 53.8 \\ 
					& CenterMask~\cite{lee2019centermask}    & ResNet-101 & 38.3 & - & - & 17.7 & 40.8 & {\bf 54.5} \\
					& BlendMask~\cite{chen2020blendmask}    & ResNet-101 & 38.4 & 60.7 & 41.3 & 18.2 & 41.5 & 53.3 \\
					& \textbf{BCNet + FCOS~\cite{tian2019fcos}}  & ResNet-101 & {\bf 39.6}    & {\bf 61.2} & {\bf 42.7} & {\bf 22.3} & {\bf 42.3} & 51.0  \\
					\hline
		\end{tabular}}}
	\end{center}
	\label{table:fully}
	\vspace{-0.2in}
\end{table*}

\begin{figure}[!t]
	\centering
	\includegraphics[width=1.0\linewidth]{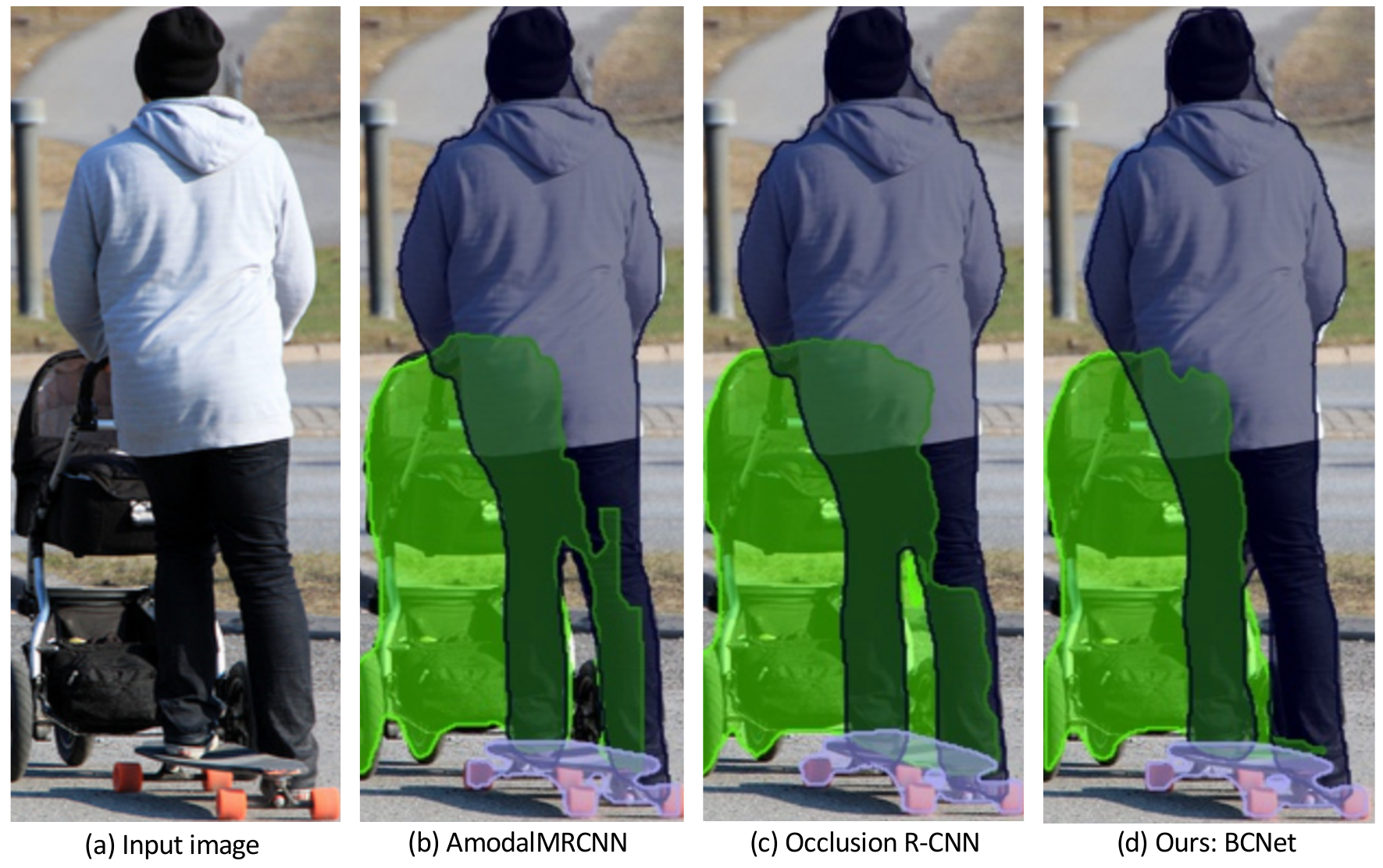}
	\caption{Qualitative results comparison of the \textbf{amodal }mask predictions on \textbf{COCOA}~\cite{zhu2017semantic} by AmodalMRCNN~\cite{follmann2019learning}, ORCNN~\cite{follmann2019learning} and our method using ResNet-50, where BCNet hallucinates a more reasonable shape for the baby carriage without producing a large portion of segmentation error. We remove the ``stuff'' background for more clarity. }
	\label{fig:amodal_example1}
	\vspace{-0.1in}
\end{figure}

\begin{figure}[!t]
	\centering
	\includegraphics[width=1.0\linewidth]{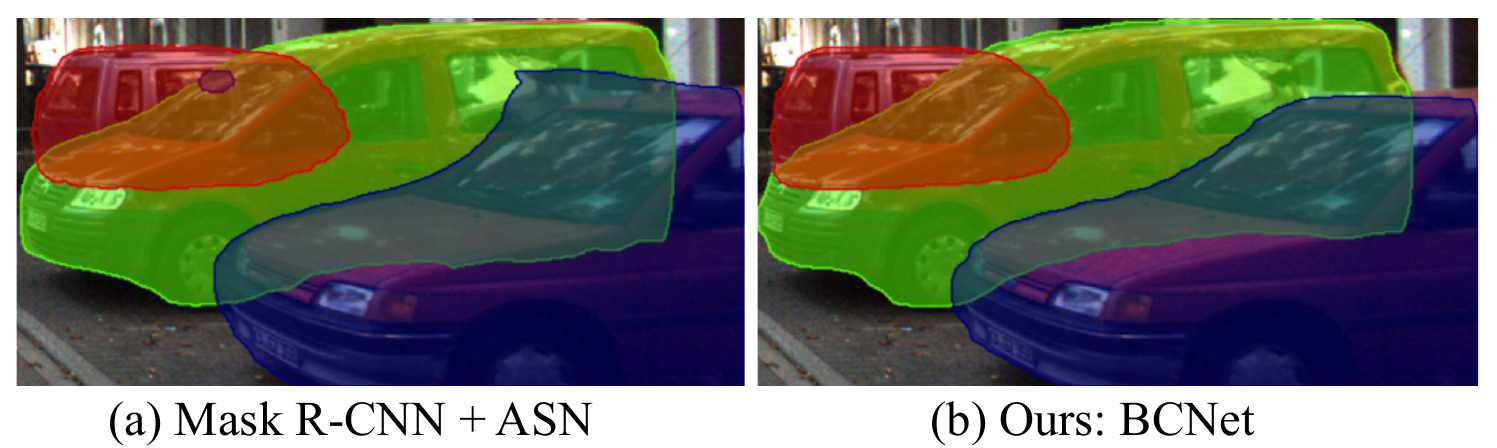}
	\caption{Qualitative results comparison of the \textbf{amodal} mask predictions on \textbf{KINS}~\cite{qi2019amodal} by Mask R-CNN + ASN~\cite{qi2019amodal} and ours, both using ResNet-101-FPN, where the boundaries of the two neighboring cars parked beside green-masked car are more reasonably estimated by BCNet. }
	\label{fig:amodal_example2}
	\vspace{-0.1in}
\end{figure}

\begin{figure*}[!h]
	\centering
	\includegraphics[width=1.0\linewidth]{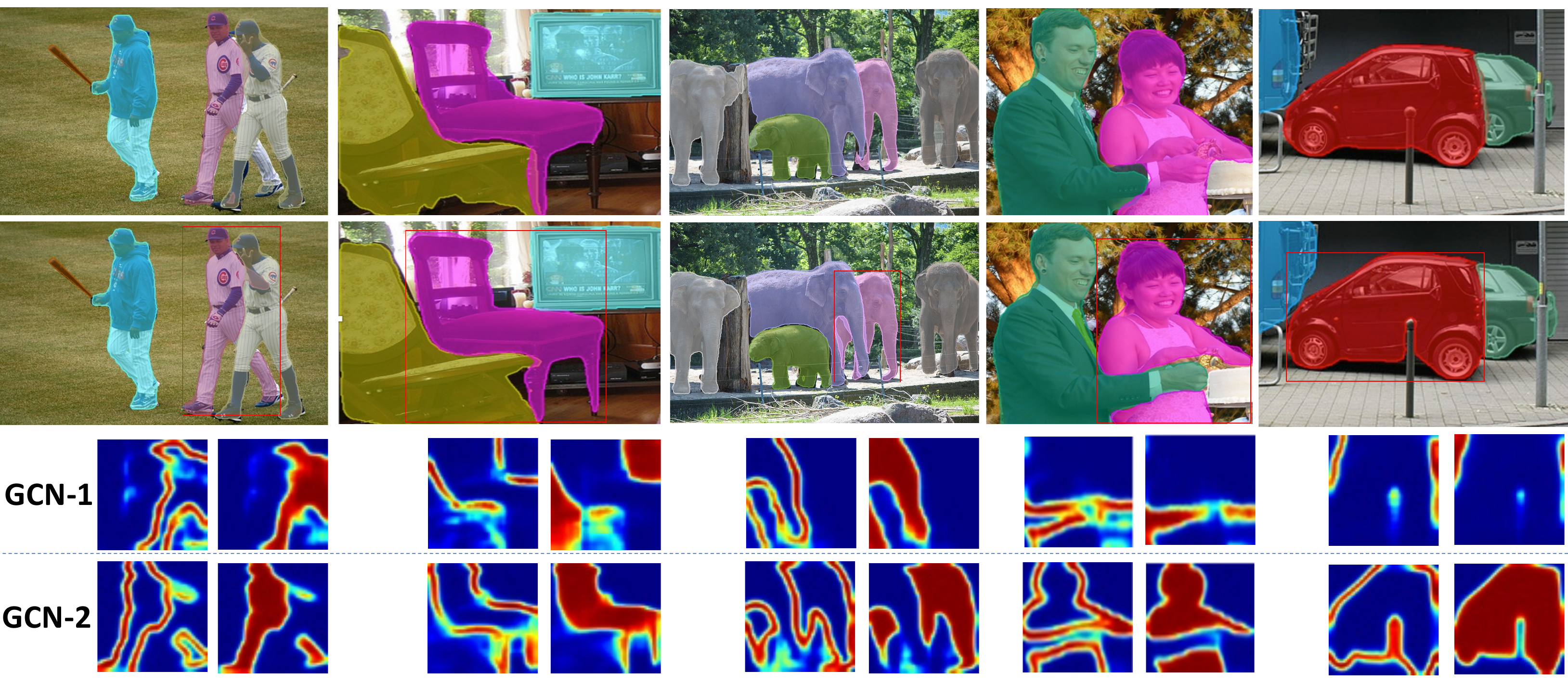}
	\vspace{-0.1in}
	\caption{Qualitative instance segmentation results of CenterMask~\cite{lee2019centermask} (top row) and our BCNet (middle row) on \textbf{COCO}~\cite{lin2014microsoft}, both using ResNet-101-FPN and FCOS detector~\cite{tian2019fcos}. The bottom row visualizes squared heatmap of contour and mask predictions by the two GCN layers for the occluder and occludee in the same~\textbf{ROI region} specified by the {\color{red} red} bounding box, which also makes the final segmentation result of BCNet more explainable 
	than previous methods. More qualitative results are available in the supplementary file.}
	\label{fig:qualitative}
	\vspace{-0.1in}
\end{figure*}

\paragraph{Comparison with Amodal Segmentation Methods}
Table~\ref{tab:cocoa} and Table~\ref{tab:kins} compare BCNet with other SOTA amodal segmentation methods on both the COCOA~\cite{zhu2017semantic} and KINS~\cite{qi2019amodal} datasets, where: 1) AmodalMask~\cite{zhu2017semantic} directly predicts amodal masks from image patches; 2) Occlusion RCNN (ORCNN)~\cite{follmann2019learning} is an extension of Mask R-CNN with both amodal and modal mask heads; 3) ASN module~\cite{qi2019amodal} contains additional occlusion classification branch and multi-level coding. 
Compared to these occlusion handling approaches, our bilayer GCN with cascaded structure still performs favorably against the state-of-the-art methods, which shows the effectiveness of BCNet in decoupling overlapping objects and mask completion under the amodal segmentation setting.
Figure~\ref{fig:amodal_example1} and Figure~\ref{fig:amodal_example2} show the qualitative comparison on COCOA and KINS respectively.
\vspace{-0.1in}

\paragraph{Evaluation on Occluded Images}
We adopt COCO-OCC split to compare the occlusion handling ability of BCNet with other methods on images with highly overlapping objects. As shown in Table~\ref{tab:occ_split}, our BCNet with Faster R-CNN detector has 31.71 \textit{AP} \textit{vs.} 30.32 for the Mask Scoring R-CNN~\cite{huang2019mask}. By further training BCNet on the synthetic occlusion dataset, the performance of~\textit{AP} and \textit{AP}$_{50}$ is significantly promoted to 32.89 and 53.25 respectively, which shows the advantage brought by this new dataset.
\vspace{-0.1in}

\paragraph{Qualitative Evaluation.}
Figure~\ref{fig:qualitative} shows 
qualitative comparison of CenterMask~\cite{lee2019centermask} and BCNet on images with overlapping objects. In each ROI region, GCN-1 detects occluding regions while GCN-2 models the partially occluded instance by directly regressing the contours and masks. For example, BCNet decouples the occluding and occluded baseball players in similar clothes into GCN-1 and GCN-2 respectively, and detects the left leg missed by CenterMask. See supplementary file for more visual comparisons.
\vspace{-0.05in}

	\section{Conclusion}
	We propose BCNet, an effective mask prediction network for addressing instance segmentation in the presence of highly-overlapping objects in two-stage instance segmentation. BCNet achieves consistent  gains on overall segmentation performance using different backbones and object detectors in both the modal and amodal settings. With explicit occluder-occludee  modeling, occluding and occluded instances are decoupled into two disjoint graph spaces, where the interaction between objects within each ROI region are explicitly considered. This effective approach will benefit future research in both occlusion handling and instance segmentation.

	{\small
		\bibliographystyle{ieee_fullname}
		\bibliography{bib}
	}
	
\end{document}